\documentclass[letterpaper, 10 pt, conference]{ieeeconf}  % Comment this line out if you need a4paper

\IEEEoverridecommandlockouts                              % This command is only needed if 
\usepackage{xcolor}

\usepackage{graphicx}

\usepackage{gensymb}
\usepackage[hyphens]{url}
\usepackage{multirow}
\usepackage{multicol}
\usepackage{tabularx}
\usepackage{dsfont}
\usepackage{url}
\usepackage{color}
\usepackage{subcaption}
\usepackage{caption}
\usepackage{booktabs}
\usepackage{dcolumn}
\usepackage{amssymb}
\usepackage{hyperref}
\usepackage{amsmath}
\usepackage{colortbl}
\usepackage{tikz}
\colorlet{lightgray}{gray!20}

\usepackage{physics}
\usepackage{graphicx}
\usepackage{float}
\usepackage{subfloat}
\usepackage{longtable}
\usepackage{threeparttable}
\usepackage{xcolor}
\usepackage{diagbox}
\usepackage{cite}
\usepackage{booktabs}
\usepackage{graphicx}
\usepackage{rotating}
\usepackage{multirow}

\usepackage{color}
\usepackage{hyperref}
\hypersetup{
    colorlinks=true,    
    linkcolor=blue,
    citecolor=royalblue,
    filecolor=magenta,      
    urlcolor=cyan,
    pdftitle={Overleaf Example},
    pdfpagemode=FullScreen,
    }

\usepackage{booktabs}
\usepackage{multirow}

 \renewcommand{\paragraph}[1]{
    \vspace{2mm}
     \noindent\textbf{#1} 
 }

\definecolor{amber(sae/ece)}{rgb}{1.0, 0.49, 0.0}

\title{\LARGE \bf
V2X-DG: Domain Generalization for Vehicle-to-Everything \\ 
Cooperative Perception}

\author{Baolu Li$^{1*}$, Zongzhe Xu$^{2*}$, Jinlong Li$^{3}$, Xinyu Liu$^{1}$, Jianwu Fang$^{4}$, Xiaopeng Li$^{5}$, Hongkai Yu$^{1\dagger}$ 
\thanks{$^{1}$Cleveland State University. $^{2}$Carnegie Mellon University. $^{3}$Texas A\&M University. $^{4}$Xi'an Jiaotong University. $^{5}$University of Wisconsin-Madison. *Equal contribution. This research is partially supported by NSF 2215388 and 2343167. $\dagger$Corresponding Author: h.yu19@csuohio.edu} 
}
\begin{document}

\maketitle
\thispagestyle{empty}
\pagestyle{empty}

%%%%%%%%%%%%%%%%%%%%%%%%%%%%%%%%%%%%%%%%%%%%%%%%%%%%%%%%%%%%%%%%%%%%%%%%%%%%%%%%
\begin{abstract}

LiDAR-based Vehicle-to-Everything (V2X) cooperative perception has demonstrated its impact on the safety and effectiveness of autonomous driving. Since current cooperative perception algorithms are trained and tested on the same dataset, the generalization ability of cooperative perception systems remains underexplored. This paper is the first work to study the Domain Generalization problem of LiDAR-based V2X cooperative perception (V2X-DG) for 3D detection based on four widely-used open source datasets: OPV2V, V2XSet, V2V4Real and DAIR-V2X. Our research seeks to sustain high performance not only within the source domain but also across other unseen domains, achieved solely through training on source domain. To this end, we propose Cooperative Mixup Augmentation based Generalization (CMAG) to improve the model generalization capability by simulating the unseen cooperation, which is designed compactly for the domain gaps in cooperative perception. Furthermore, we propose a constraint for the regularization of the robust generalized feature representation learning: Cooperation Feature Consistency (CFC), which aligns the intermediately fused features of the generalized cooperation by CMAG and the early fused features of the original cooperation in source domain. Extensive experiments demonstrate that our approach achieves significant performance gains when generalizing to other unseen datasets while it also maintains strong performance on the source dataset.
\end{abstract}

\section{Introduction}
\label{sec:introduction}
Precisely perceiving the surrounding environment is crucial for the safety of autonomous driving~\cite{caesar2020nuscenes}. Recently with the LiDAR sensor, multi-vehicle based cooperative perception~\cite{li2022v2x,chen2019f} has demonstrated its superior 3D object detection over single-vehicle based perception by utilizing Vehicle-to-Everything (V2X) communication to share information between multiple agents, \textit{i.e.}, Connected Autonomous Vehicles (CAVs) and smart infrastructures, thereby achieving a broader perception range and more precise sensing accuracy. With the availability of numerous V2X cooperative perception datasets and benchmarks~\cite{xu2022opv2v,xu2022v2x,li2022v2x,xu2023v2v4real,yu2022dair,liu2024v2x} that have been collected and publicized, many methods~\cite{xu2022cobevt,lu2023robust} have demonstrated their effectiveness. However, those methods assume that the training data and deployed testing scenarios belong to the same data distribution, leaving the generalization ability of V2X cooperative perception as an open question.

Empirically, there will be a performance drop when deploying a trained model to unseen domains, and this phenomenon is because of the domain gap. In this paper, we firstly study the Domain Generalization (DG) problem of the LiDAR based V2X cooperative perception system for 3D detection. After carefully investigating the widely-used open source LiDAR based V2X datasets (OPV2V~\cite{xu2022opv2v}, V2XSet~\cite{xu2022v2x}, V2V4Real~\cite{xu2023v2v4real}, DAIR-V2X~\cite{yu2022dair}) in this research area, we observe that the domain gap in cooperative perception includes the following two perspectives.

\begin{figure}[!t]
\centering
\subfloat{ 
\includegraphics[width=1\columnwidth]{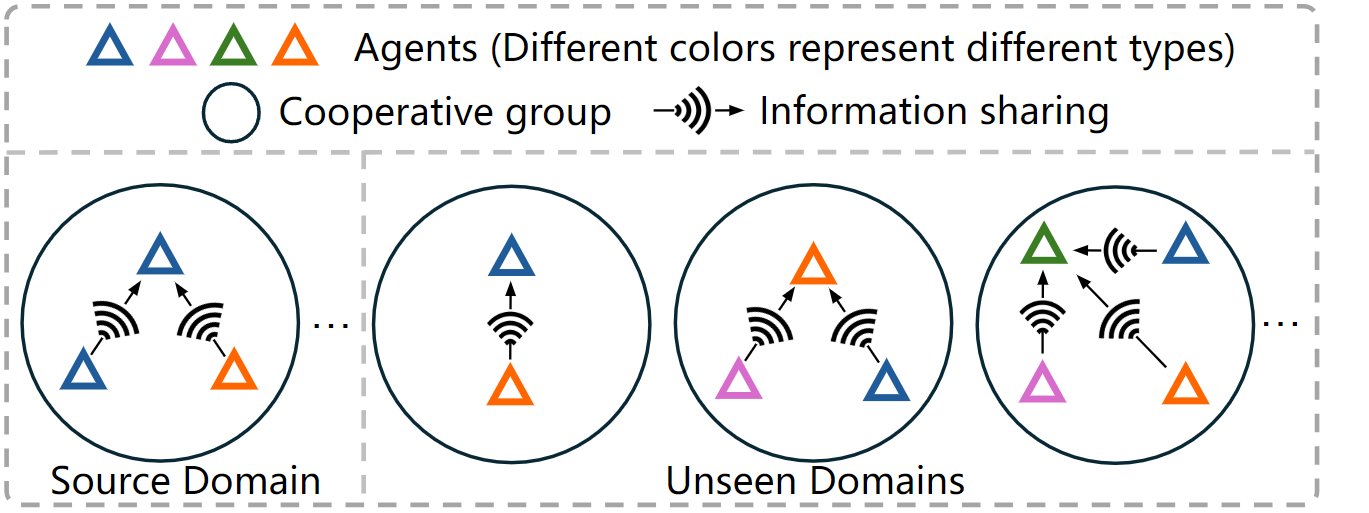}%
}
\caption{Illustration of the domain gap (\textit{Agent Number Distribution Gap}, \textit{Cooperative Agent Type Gap}) for V2X cooperative perception between source domain to unseen domains. Best viewed in color.}
\vspace{-1.em}
\label{fig:gap}
\end{figure}

\begin{itemize}
    \item \textbf{Agent Number Distribution Gap:}
    As shown in Fig.~\ref{fig:gap}, the agent number distribution between the source domain and unseen domains might differ. This variation in agent numbers leads to a diverse pool of shared visual information, which can impact the generalization ability when the trained model is deployed in scenarios with varying agent numbers.

    \item \textbf{Cooperative Agent Type Gap:} 
    As shown in Fig.~\ref{fig:gap}, the agent types within cooperative groups in the source domain might be different from those in the unseen domains. This variation in cooperative agent types leads to discrepancies in the cooperative perception, as the data aggregation from different agent types might introduce inconsistencies.
\end{itemize}

As shown in the left part of Table~\ref{tab:tab1}, the agent number distribution across the four open source LiDAR based  datasets varies, particularly in V2V4Real and DAIR-V2X, where scenarios with only one or two agents are presented. The right part of Table~\ref{tab:tab1} illustrates the cooperative agent types in different datasets. The agent types are categorized into A, B, C, D, and E based on their LiDAR Beams, capturing range, vertical Field-of-View (FOV), error, and Simulated or Real, as detailed in Table~\ref{tab:tab2}. These observations underscore the critical need for advancing methodologies of domain generalization in V2X cooperative perception.

\begin{table*}[htb]
\caption{Agent number distribution (probability) and cooperative agent type of V2X cooperative perception datasets.}
\centering
\resizebox{0.85\textwidth}{!}{%
\begin{tabular}{c|ccccc|ccccc}
\toprule
\multirow{2}{*}{Dataset} & \multicolumn{5}{c|}{Agent Number Distribution (probability)}                                                                                                                                             & \multicolumn{5}{c}{Cooperative Agent Type}                                                                                                \\
                         & 1                                   & 2                                   & 3                                   & 4                                   & 5                                  & A                         & B                         & C                         & D                         & E                         \\ \midrule
OPV2V~\cite{xu2022opv2v}                    & \checkmark (7.87\%)  & \checkmark (48.46\%) & \checkmark (26.57\%) & \checkmark (16.20\%) & \checkmark (0.90\%) & \checkmark &                           &                           &                           &                           \\
V2XSet~\cite{xu2022v2x}                   & \checkmark (12.75\%) & \checkmark (39.00\%) & \checkmark (33.15\%) & \checkmark (13.41\%) & \checkmark (1.69\%) & \checkmark & \checkmark &                           &                           &                           \\
V2V4Real~\cite{xu2023v2v4real}                 & \checkmark (9.80\%)  & \checkmark (90.20\%) &  -                                   &   -                                 &     -                               &                           &                           & \checkmark &                           &                           \\
DAIR-V2X~\cite{yu2022dair}                 & \checkmark (9.20\%)  & \checkmark (90.80\%) &  -                                   &   -                                 &   -                                 &                           &                           &                           & \checkmark & \checkmark \\ \bottomrule
\end{tabular}}
\label{tab:tab1}
\vspace{-1.5em}
\end{table*}

\begin{table}[htb]
\caption{Detailed LiDAR setup for different agent types.}
\centering
\resizebox{1\columnwidth}{!}{%
\begin{tabular}{c|cccccc}
\toprule
\multirow{2}{*}{Type} & LiDAR & Range & FOV                      & Error               & \multirow{2}{*}{Real/Sim} & \multirow{2}{*}{Class} \\
                      & Beams & (m)   & (\degree) & (cm)                &                           \\ \midrule
A                     & 64    & 120   & [-25, 5]                   & $\pm2$ & Sim & Vehicle                      \\
B                     & 32    & 120   & [-25, 5]                   & $\pm2$ & Sim & Infrastructure                    \\
C                     & 32    & 200   & [-25, 15]                  & $\pm3$ & Real & Vehicle                     \\
D                     & 40    & 200   & [-30, 10]                  & -                   & Real & Vehicle                     \\
E                     & 300   & 280   & [-30, 10]                  & $\pm3$ & Real & Infrastructure                     \\ \bottomrule
\end{tabular}}
\label{tab:tab2}
\vspace{-1.5em}
\end{table}

In this paper, we propose the first domain generalization approach for V2X cooperative perception system, named as \textbf{V2X-DG}, which aims to not only increase performance on unseen domains but also maintain the good performance on source domain. To bridge the domain gap in V2X field, we propose the novel Cooperative Mixup Augmentation based Generalization (CMAG) to simulate the unseen cooperative data. Specifically, we manipulate point clouds from source collaborative agents (vehicles or infrastructures) to form mixup agents, by utilizing our two precisely designed components: Point Augmentation, Probabilistic Gate. In this way, the mixup agent shares the ground truth labels with the source data while also mimicking the unseen cooperative data. Furthermore, we propose a new constraint for the regularization of the generalized feature representation learning: Cooperation Feature Consistency (CFC), which aligns the intermediately fused features of the generalized cooperation by CMAG and the early fused features of the original cooperation in source domain. The proposed CFC will help the model learn robust generalizable feature representation.

Additionally, we establish standardized training and evaluation protocols for V2X-DG, addressing the lack of setups for domain generalization. All V2X-DG experiments are conducted on four 
widely-used public  datasets~\cite{xu2022opv2v,xu2022v2x,xu2023v2v4real,yu2022dair}. Our method demonstrates superior performance compared to the baseline and comparison methods. The contributions of this paper are summarized as follows.

\begin{itemize}
    \item To the best of our knowledge, we propose the \textbf{first research} of domain generalization for LiDAR-based V2X cooperative perception (3D detection) to promote the model performance on unseen domains, named as V2X-DG.

    \item We propose a novel Cooperative Mixup Augmentation based Generalization (CMAG) to enhance the model generalization by  effectively simulating domain gaps in V2X cooperative perception and a new Cooperation Feature Consistency (CFC) as regularization for the robust generalized feature representation learning.

    \item We evaluate the proposed method on four widely-used public LiDAR-based V2X cooperative perception datasets (OPV2V, V2XSet, V2V4Real, DAIR-V2X), and the experimental results demonstrate the significant performance improvement on unseen domains and very good performance maintenance on source domain. 
\end{itemize}

\section{Related Work}\label{Sec:Related_Work}

\noindent \textbf{Cooperative Perception.}
While single-agent perception systems face challenges of occlusions and short-range detection, the cooperative systems mitigate these issues through various inter-agent communication strategies. Given bandwidth constraints, researchers have proposed many methods of intermediate fusion (sharing hidden layer outputs between agents), receiving significant attention for its balance between prediction accuracy and communication bandwidth~\cite{xu2022v2x,huwhere2comm,chen2019f}. Numerous efforts are made to develop different fusion techniques under diverse scenarios~\cite{li2023s2r,li2024Break,li2024advattack}. For instance, V2VNet~\cite{wang2020v2vnet} employs a graph neural network to facilitate feature fusion among agents, while When2com~\cite{huwhere2comm} introduces a spatial confidence-aware communication strategy to enhance efficiency. SyncNet~\cite{lei2022latency} addresses time synchronization by compensating for communication delays, and AttFuse~\cite{xu2022opv2v} leverages attention mechanisms to improve fusion of intermediate features. Additionally, transformer-based architectures, such as V2X-ViT~\cite{xu2022v2x} and CoBEVT~\cite{xu2022cobevt}, have been tailored to this domain, demonstrating outstanding performance. However, most V2X cooperative perception methods are limited to experiments on the same dataset, leaving their generalization ability unexplored. This paper is the first work to investigate the domain generalization problem of LiDAR-based V2X cooperative perception (V2X-DG), aiming to enhance performance on unseen domains and maintain strong performance on source domain.

\noindent \textbf{Domain Generalization for Perception.}
Domain generalization (DG) is a critical challenge in the field of perception, particularly when systems maintains high performance in unseen environments despite being trained exclusively on source domains. While DG has been extensively studied in 2D and 3D vision tasks like object detection~\cite{vidit2023clip}, semantic segmentation~\cite{kim2023single}, its application to V2X cooperative perception systems, especially those relying on 3D point clouds from LiDAR, remains underexplored. Existing strategies for DG, such as disentangling domain-specific and domain-invariant features~\cite{liu2023d2ifln, bui2021exploiting}, aligning source domain distributions~\cite{chen2023domain, matsuura2020domain}, and employing domain augmentation techniques~\cite{zhou2020learning}, have proven effective in various contexts but have not been fully adapted to the V2X cooperative perception domain. Especially, the agent number distribution gap and cooperative agent type gap in different domains introduce more complexities and prevent the direct application of traditional domain generation methods. This paper aims to provide a comprehensive examination of such challenges and introduces a unified training schema that mitigate all these gaps at the same time.

\section{Methodology}\label{Sec:Method}

\subsection{Overview of V2X-DG Architecture}
\begin{figure*}[!t]
\centering
\subfloat{ 
\includegraphics[width=1\textwidth]{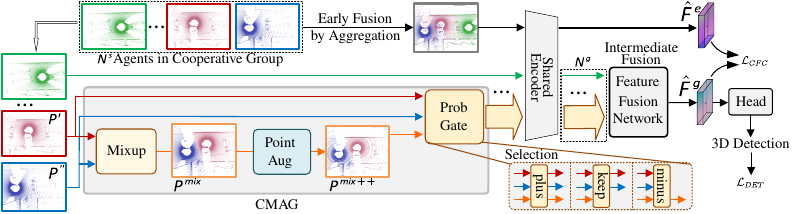}%
}
\caption{\textbf{Overview V2X-DG framework.} CMAG bridges the Agent Number Distribution Gap and Cooperative Agent Type Gap by three components: Mixup Agent, Point Augmentation, and Probabilistic Gate.  $\mathcal{L}_{CFC}$ is the  regularization for robust generalized feature representation learning. V2X-DG is only used in the source-domain  training. Best view in color.}
\label{fig:overall}
\vspace{-1.5em}
\end{figure*}

Our objective is to develop a LiDAR-based V2X cooperative perception model that demonstrates robust performance on unseen domains, even when it is trained exclusively on source domain. Given a spatial group of $N^s$ agents (\textit{i.e.}, vehicles or infrastructures) within the communication range in source domain for training, all point clouds are projected to the ego-agent's coordinate system. To bridge cooperative gaps, we firstly deploy Cooperative Mixup Augmentation based Generalization (CMAG) to the source cooperative group and obtain the generalized cooperative group  $\mathbf{P}^{g} = \{ \mathbf{p}^{i} \}^{N^{g}}_{i=1} $, where $N^{g}$ is the agent number after CMAG, $\mathbf{p}^{i}$ is the point cloud of $i$-th agent. Then, all agents in $\mathbf{P}^{g}$ go through a shared Encoder $\mathbf{E}(\cdot)$ for feature extraction and their intermediate features $\mathbf{F}^{g}= \{ \mathbf{f}^{i}  \}^{N^{g}}_{i=1} $ are acquired. After feature sharing between ego-agent and other-agents via V2X communication, the ego-agent will get all the intermediate features in $\mathbf{P}^{g}$ and fuse them by a Feature Fusion Network. Finally, the fused features $\hat{\textbf{F}}^g$ are fed into a prediction header $\mathbf{H}$ for 3D bounding-box regression and classification. We also introduce Cooperation Feature Consistency (CFC) as regularization to promote robust   generalizable representation learning. The 3D object detection performance of the cooperative perception model will be assessed on both the source domain and unseen domains. Fig.~\ref{fig:overall} shows the proposed V2X-DG framework, and the details of each part will be explained in the following sections.

\subsection{Cooperative Mixup Augmentation based Generalization}
To bridge the domain gap in cooperative perception, we propose Cooperative Mixup Augmentation based Generalization (CMAG) to simulate unseen cooperative data. In order to enhance generalization while maintaining performance within the source domain, we let  the model simultaneously learn the existing cooperative data in the source domain and generalized cooperative data. Specifically, we introduce a mixup agent based on existing agents. Since the mixup agent originates from the source domain, introducing it into the collaborative process will not harm the  performance within source domain. In addition, we design a novel \textit{Point Augmentation} to reduce the \textit{Cooperative Agent Type Gap} and introduce a new \textit{Probabilistic Gate} to reduce \textit{Agent Number Distribution Gap}. These modules are executed dynamically during each training iteration. The CMAG is employed solely for training and is discarded in testing.

\noindent \textbf{Mixup Agent.}
To avoid harms to the source domain, we propose a mixup agent as an augmentation strategy from the existing cooperative data in source domain. Specifically, the point clouds of the two  nearest agents in the source-domain cooperative group are chosen as $\mathbf{p'}$ and $\mathbf{p''}$ firstly. Then, their center $\mathbf{c'}$ and $\mathbf{c''}$ are obtained based on the bird's eye view (x-y plane). After that, we get the normal vector between $\mathbf{c'}$ and $\mathbf{c''}$ and apply random rotation to the normal vector, leading to a split vector. The points in $\mathbf{p'}$ and $\mathbf{p''}$ are individually cut based on the split vector, with $\mathbf{p'_{side}}$ retaining the points on one side of the split vector and $\mathbf{p''_{side}}$ retaining the  points on the opposite side. Finally, the $\mathbf{p}^{mix}$ is their combination, which  can be expressed as:

\begin{equation}
    \mathbf{p}^{mix} = \mathbf{p'_{side}} \cup \mathbf{p''_{side}}.
\end{equation}

Because the mixup agent is derived from the combination of agents in the source domain, its introduction does not affect the ground truth of the following 3D detection task.

\noindent \textbf{Point Augmentation.}
The Point Augmentation (PA) aims to apply several augmentations to the mixup agent in the point cloud field, which bridges the \textit{Cooperative Agent Type Gap} as mentioned in   Section~\ref{sec:introduction}. We focus on two aspects: LiDAR density and LiDAR setup, which have been illustrated at Table~\ref{tab:tab2}.

The agent LiDAR beams in unseen domains are diverse, making point density augmentation quite valuable. Following~\cite{milioto2019rangenet++}, we project every point $[x_i, y_i, z_i]$ in $\mathbf{p}^{mix}$ into a range-view image:      

\begin{equation}
\left( r_i^x, r_i^y \right) = \left( 
 \frac{1}{2} \left[1 - \frac{\theta_i}{\pi}\right] W ,
\left[1 - \frac{\varphi_i + f_{\text{max}}}{f}\right] H 
\right),
\end{equation}
where $r_i^x, r_i^y$ denote the coordinates in the range-view image, $\theta_i = \arctan(y_i, x_i)$ is the azimuth, $\varphi_i = \arctan(z_i, \sqrt{(x_i)^2 + (y_i)^2})$ is the altitude, $f_{\text{max}}$, $f$ denote the upper bound FOV and FOV itself, and $H$, $W$ are the height and width of the image. The height $H$ corresponds to the LiDAR beams after projection. The PA for LiDAR density includes downsampling and upsampling of LiDAR beams, which correspond to the removal and interpolation beams along the $H$-dimension, respectively. We implement downsampling or upsampling randomly to the mixup agent $\mathbf{p}^{mix}$, leading to  $\mathbf{p}^{mix+}$ for the generalized LiDAR densities.

The agent LiDAR setup is also uncertain in unseen domains, with diverse parameters such as capturing range, vertical FOV, error, and real/simulated data, as shown in Table~\ref{tab:tab2}. To address these uncertainties, PA for LiDAR setup is further  designed by mimicking various LiDAR setups. Here, we deploy three random augmentations: random rotting $\Gamma (\cdot)$, arbitrary scaling $\Delta (\cdot)$, noise translation $\Theta (\cdot)$. Those augmentations are performed on $\mathbf{p}^{mix+}$ as:

\begin{equation}
    \mathbf{p}^{mix++} = \Theta(\Delta(\Gamma 
 (\mathbf{p}^{mix+}))),  
\end{equation}
leading to $\mathbf{p}^{mix++}$ for the generalized LiDAR setups.

\noindent \textbf{Probabilistic Gate.}
Within the same domain, the distribution of agent number is almost  identical between training set and testing set. However, the agent number distribution gap becomes apparent when considering the distribution across multiple domains. As shown in Table~\ref{tab:tab1}, the agent number follows different distributions across various datasets. We then   conduct a statistical analysis across all datasets in Table~\ref{tab:tab1} to estimate the comprehensive distribution of the agent number, which is a much more generalizable distribution. Our Probabilistic Gate aims to make the cooperative agent number distribution in source domain more close to the comprehensive distribution.

Let us define $\Phi_s(x)$ as the weight (probability) of the agent number $x$ in the source domain distribution and $\Phi_c(x)$ as the weight of the agent number $x$ in the comprehensive distribution. We have three rules to adjust the number of agents in each training  iteration. 1) If $\Phi_s(x)$ is greater than $\Phi_c(x)$, the agent number $x$ has a larger weight in source domain, so we reduce the weight of the agent number $x$. 2) If they are the same, we keep them unchanged. 3) Otherwise, we  increase the weight of the agent number $x$.

Given two nearest agents in a cooperative group, we then implement the above rules by three gates. 1) Plus Gate: the mixup agent $\mathbf{p}^{mix}$ could be directly added to the cooperative group, making the agent number plus one. 2) Keep Gate: we keep the agent number unchanged. 3) Minus Gate: Otherwise, we can remove both $\mathbf{p'}$ and $\mathbf{p''}$ to use $\mathbf{p}^{mix}$ only, making the agent number minus one. This gate selection is implemented by computing the corresponding responses and likelihoods:

\begin{equation}
    r_{+} = \text{max}(0,\frac{\Phi_c(N^s+1) - \Phi_s(N^s+1) }{\text{max}(\Phi_s(N^s+1),\epsilon)}),  
\end{equation}

\begin{equation}
    r_{-} = \text{max}(0,\frac{\Phi_c(N^s-1) - \Phi_s(N^s-1) }{\text{max}(\Phi_s(N^s-1),\epsilon)}),  
\end{equation}
where $\epsilon$ is a small constant used to prevent division by zero, and the response of keeping  agent number unchanged $r_=$ is set to 1, and then we normalize $r_{+}$, $r_{-}$, $r_{=}$ to obtain the their likelihoods. Finally, we use these likelihoods to guide the gate selection during each training iteration.

\subsection{Cooperation Feature Consistency}
By the data manipulation of the proposed CMAG, the model will learn the generalizable feature representation for V2X cooperative perception. To avoid overfitting and enhance generalization, we propose a new constraint as the regularization to learn the robust generalizable feature representation, which is named as Cooperation Feature Consistency (CFC) in this paper.

Given the original cooperative group of agents in source domain, we simply aggregate all point clouds through early fusion as~\cite{li2024di,li2021learning}. Then, the aggregated point clouds of all agents will be fed to the shared Encoder $\mathbf{E}(\cdot)$ and get the early fused feature $\hat{\mathbf{F}}^e$. After passing the CMAG and Feature Fusion Network, the intermediately fused feature of the generalized cooperation $\hat{\mathbf{F}}^g$ can be obtained. Using the early fused feature as guidance, our CFC aligns $\hat{\mathbf{F}}^g$ and $\hat{\mathbf{F}}^e$ by designing the following $L_1$ loss function as regularization in training: 

\begin{equation}
\mathcal{L}_{CFC} = ||  \hat{\mathbf{F}}^{g} -  \hat{\mathbf{F}}^{e} ||_1.
\end{equation}

\subsection{Overall Loss Function}
The focal loss~\cite{lin2017focal} and smooth $L_1$ loss are used as the 3D object detection loss $\mathcal{L}_{DET}$~\cite{xu2023v2v4real} in the training process. The final total loss is the combination of the detection loss  and the Cooperation Feature Consistency loss $\mathcal{L}_{CFC}$ as 

\begin{equation}
    \mathcal{L}_{Total} = w_1 \mathcal{L}_{DET} + w_2 \mathcal{L}_{CFC},
   \label{eq:total_loss}
\end{equation}
where $w_1$ and $w_2$ are the balance weights.

\section{Experiments}\label{Sec:Experiment}

\subsection{Datasets}

We utilize four widely-used public LiDAR based V2X cooperative perception datasets to train and evaluate the generalization performance of 3D object detection. 

\noindent \textbf{OPV2V}~\cite{xu2022opv2v} 
is a simulated dataset designed for Vehicle-to-Vehicle (V2V) communication, providing synchronized LiDAR from multiple vehicles to simulate real-world driving environments. It has   6,764/1,981/2,719 (train/validation/test) frames of point cloud collected from eight simulated town in CARLA~\cite{dosovitskiy2017carla} and OpenCDA~\cite{10045043}.

\noindent \textbf{V2XSet}~\cite{xu2022v2x} 
is a simulated dataset designed for V2X communication, consisting of LiDAR data from both vehicles and smart infrastructures. It contains 6,694/1,920/2,833 (train/validation/test) frames of point clouds, which is also collected in CARLA~\cite{dosovitskiy2017carla} and OpenCDA~\cite{10045043}.

\noindent \textbf{V2V4Real}~\cite{xu2023v2v4real} is a real-world large-scale V2V cooperative perception dataset collected with two CAVs in Columbus, OH, USA under abundant driving scenarios. It contains 14,210/2,000/3,986 (train/validation/test) frames.

\noindent \textbf{DAIR-V2X}~\cite{yu2022dair} is a real-world large-scale  V2X dataset that focuses on Vehicle-to-Infrastructure (V2I) communication collected in Beijing, China from 28 distinct intersections. It contains 35,627/14,251/21,376 (train/validation/test) frames.

\subsection{Experimental Setup}

\noindent \textbf{Evaluation Metrics.} 
We measure performances of 3D detection result with Average Precision (AP) over two different Intersection-over-Union (IoU) threshold: $0.3$ and $0.5$. To assess the generalization performance across both the source and unseen domains, we also calculate the mean value of the APs across all datasets. Since the evaluation range and V2X communication range differ across the four datasets, we follow their default respective configurations~\cite{xu2022opv2v,xu2022v2x,xu2023v2v4real,yu2022dair}.

\noindent \textbf{Experimental Settings.}
Our goal is to enhance the perception performance of V2X cooperative perception systems by applying domain generalization (DG) techniques to improve detection in unseen domains. We evaluate detection performance across four benchmark datasets: OPV2V~\cite{xu2022opv2v}, V2XSet~\cite{xu2022v2x}, V2V4Real~\cite{xu2023v2v4real}, and DAIR-V2X~\cite{yu2022dair}. All methods use the same cooperative perception approach, AttFuse~\cite{xu2022opv2v} (implemented as Feature Fusion Network), with a PointPillar~\cite{lang2019pointpillars}-based Encoder for consistency in comparison. 
We assess all methods under four key settings: \textbf{OPV2V2ALL}, \textbf{V2XSet2ALL}, \textbf{V2V4Real2ALL}, and \textbf{DAIR-V2X2ALL}. In each scenario, all DG methods are trained exclusively on the training set of one dataset and then tested across the testing sets of all the 
 four datasets.

\subsection{Comparison Methods.}
The Domain Generalization (DG) of LiDAR-based cooperative perception is an unexplored problem, so we introduce some general DG methods to this area.

\noindent \textbf{Baseline}: In~\cite{xu2022opv2v}, AttFuse is used as Feature Fusion Network for cooperative perception, which is considered as baseline here. Baseline uses the source domain for training and is directly evaluated on the target domains.

\noindent \textbf{IBN-Net}: In~\cite{pan2018two},  
IBN-Net merges Instance Normalization (IN) with Batch Normalization (BN) to keep the content information and learn appearance invariant features. Out of several versions of the IBN block, we follow the IBN-b block~\cite{pan2018two}, due to its great performance, by directly adding IN after the first, second, and third Conv layer of the Encoder.

\noindent \textbf{MLDG}: In~\cite{li2018learning}, MLDG is a meta-learning based DG method, which splits the training batch data into a meta-train set and meta-test set to improve the generalization ability. Following~\cite{kim2023single}, we implement two versions based on different splitting strategies: MLDG$_a$ and MLDG$_b$. MLDG$_a$ randomly splits the training batch data into two equal number subsets, which are meta-train set and meta-test set. In contrast, MLDG$_b$ selects one sample for the meta-train set first and then calculates the cosine distance to the other sample's fused feature maps. The sample with the closest distance is added to the meta-train set, while the remaining sample with further distance will be assigned to the meta-test set.

\begin{figure*}[!ht]
\centering
\footnotesize
\def\xwidth{0.24}
\def\yheight{0.14}
\def\xem{-2pt}
\def\im_shift{0.01\textwidth}
\setlength{\tabcolsep}{0.5pt}
\begin{tabular}{ccccc}
& OPV2V~\cite{xu2022opv2v} & V2XSet~\cite{xu2022v2x} & V2V4Real~\cite{xu2023v2v4real} & DAIR-V2X~\cite{yu2022dair}\\
\multirow[t]{1}{*}[\im_shift]{\begin{sideways}  Baseline~\cite{xu2022opv2v}  \end{sideways}} &
\includegraphics[width=\xwidth\textwidth,height=\yheight\textwidth]{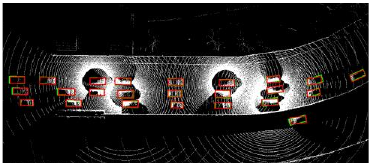}
& \includegraphics[width=\xwidth\textwidth,height=\yheight\textwidth]{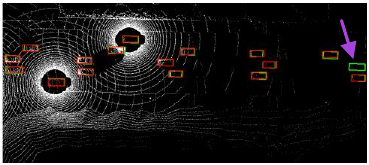}
& \includegraphics[width=\xwidth\textwidth,height=\yheight\textwidth]{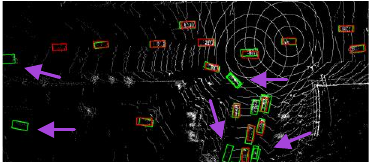}
& \includegraphics[width=\xwidth\textwidth,height=\yheight\textwidth]{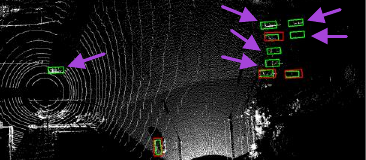} \\
\multirow[t]{1}{*}[\im_shift]{\begin{sideways}  IBN-Net~\cite{pan2018two} \end{sideways}} &
\includegraphics[width=\xwidth\textwidth,height=\yheight\textwidth]{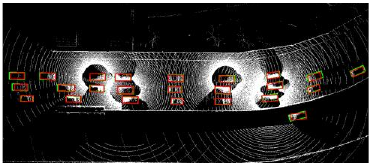}
& \includegraphics[width=\xwidth\textwidth,height=\yheight\textwidth]{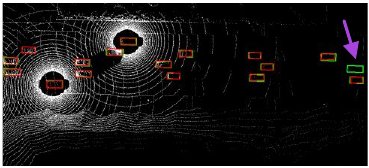} 
& \includegraphics[width=\xwidth\textwidth,height=\yheight\textwidth]{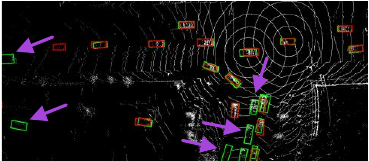}
& \includegraphics[width=\xwidth\textwidth,height=\yheight\textwidth]{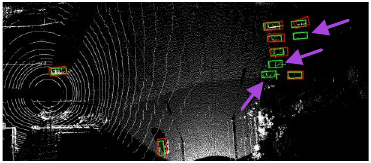} \\
\multirow[t]{1}{*}[\im_shift]{\begin{sideways} V2X-DG (Ours) \end{sideways}} &
\includegraphics[width=\xwidth\linewidth,height=\yheight\textwidth]{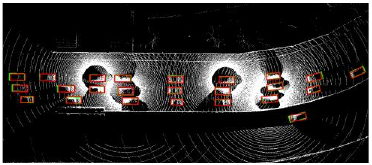}
& \includegraphics[width=\xwidth\linewidth,height=\yheight\textwidth]{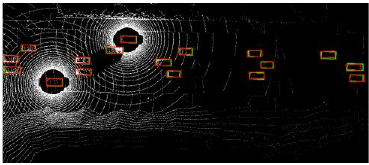}
& \includegraphics[width=\xwidth\linewidth,height=\yheight\textwidth]{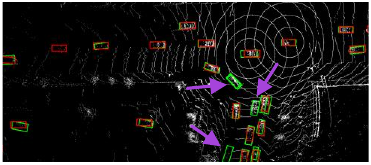}
& \includegraphics[width=\xwidth\linewidth,height=\yheight\textwidth]{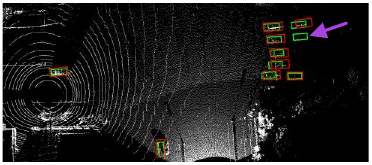} \\
\end{tabular}
\caption{\textbf{Qualitative comparison on OPV2V2ALL setting.} \textcolor{green}{Green} and \textcolor{red}{red} 3D bounding boxes represent the ground truth and prediction respectively. The  detection errors are highlighted using \textcolor{violet}{purple} arrows. OPV2V is used as source domain here.}
\label{fig:det_vis}
\vspace{-1.em}
\end{figure*}

\subsection{Quantitative Comparisons.}

\begin{table}[htb]
\centering
\caption{\textbf{OPV2V2ALL setting.} The results are reported by training on the source domain OPV2V only.}
\resizebox{1.\columnwidth}{!}{%
\begin{tabular}{c|c|ccc|c}
\toprule
\multirow{2}{*}{Method} & OPV2V      & V2XSet     & V2V4Real   & DAIR-V2X   & Mean         \\
                        & AP@0.3/0.5 & AP@0.3/0.5 & AP@0.3/0.5 & AP@0.3/0.5 & AP@0.3/0.5 \\ \midrule
Baseline                    & 90.57/90.25  & 85.13/84.39  & 40.96/38.74  & 41.44/36.79  & 64.27/62.54  \\
IBN-Net                 & 92.71/92.29  & 89.58/\textbf{88.59}  & 43.65/41.73  & 38.93/32.53  & 66.22/63.79  \\
MLDG$_a$                 & \textbf{92.73}/\textbf{92.31}  & \textbf{89.65}/88.56  & 41.91/39.68  & 40.85/36.21  & 66.29/64.19  \\
MLDG$_b$                   & 90.12/89.56 & 84.51/83.26 & 38.45/34.81 & 34.32/29.99 & 61.85/59.40  \\
Ours                    & 92.69/92.03 & 88.48/87.42 & \textbf{46.11}/\textbf{44.17} & \textbf{46.08}/\textbf{41.13} & \textbf{68.34}/\textbf{66.19}  \\ \bottomrule
\end{tabular}}
\label{tab:opv2v}
\end{table}

\begin{table}[htb]
\centering
\caption{\textbf{V2XSet2ALL setting.} The results are reported by training on the source domain V2XSet only.}
\resizebox{1.\columnwidth}{!}{%
\begin{tabular}{c|c|ccc|c}
\toprule
\multirow{2}{*}{Method} & V2XSet      & OPV2V     & V2V4Real   & DAIR-V2X   & Mean         \\
                        & AP@0.3/0.5 & AP@0.3/0.5 & AP@0.3/0.5 & AP@0.3/0.5 & AP@0.3/0.5 \\ \midrule
Baseline                    & 88.56/87.19 & 91.51/90.66 & 39.50/37.83 & 39.20/34.41 & 64.70/62.52  \\
IBN-Net                 & 88.99/87.70 & 91.67/90.99 & 42.01/39.90 & 28.83/23.97 & 62.87/60.64  \\
MLDG$_a$                 & 89.19/\textbf{88.08} & \textbf{92.00}/\textbf{91.27} & 40.15/38.31 & 36.94/32.45 & 64.57/62.53  \\
MLDG$_b$                   & 87.06/85.96 & 89.83/88.98 & 41.93/38.29 & 35.62/30.69 & 63.61/60.98  \\
Ours                    & \textbf{89.24}/87.34 & 91.74/90.63 & \textbf{43.73}/\textbf{40.66} & \textbf{44.26}/\textbf{38.80} & \textbf{67.24}/\textbf{64.36}  \\ \bottomrule
\end{tabular}}
\label{tab:v2xset}
\end{table}

\begin{table}[htb]
\centering
\caption{\textbf{V2V4Real2ALL setting.} The results are reported by training on the source domain V2V4Real only.}
\resizebox{1.\columnwidth}{!}{%
\begin{tabular}{c|c|ccc|c}
\toprule
\multirow{2}{*}{Method} & V2V4Real      & OPV2V     & V2XSet   & DAIR-V2X   & Mean         \\
                        & AP@0.3/0.5 & AP@0.3/0.5 & AP@0.3/0.5 & AP@0.3/0.5 & AP@0.3/0.5 \\ \midrule
Baseline                    & 53.48/51.10 & 48.80/46.99 & 45.81/44.64 & 44.51/39.72 & 48.15/45.61  \\
IBN-Net                 & 54.94/51.45 & 47.54/44.70 & 41.27/39.37 & \textbf{47.42}/\textbf{41.23} & 47.79/44.19  \\
MLDG$_a$                 & 54.24/51.89 & 51.94/49.56 & 47.92/46.12 & 46.05/41.05 & 50.04/47.16  \\
MLDG$_b$                   & 53.17/51.01 & 44.41/42.66 & 42.43/41.15 & 41.73/37.43 & 45.44/43.06  \\
Ours                    & \textbf{56.07}/\textbf{53.99} & \textbf{53.91}/\textbf{51.96} & \textbf{50.29}/\textbf{48.84} & 45.36/40.77 & \textbf{51.41}/\textbf{48.89}  \\ \bottomrule
\end{tabular}}
\label{tab:v2v4real}
\end{table}

\begin{table}[htb]
\centering
\caption{\textbf{DAIR-V2X2ALL setting.} The results are reported by training on the source domain DAIR-V2X only.}
\resizebox{1.\columnwidth}{!}{%
\begin{tabular}{c|c|ccc|c}
\toprule
\multirow{2}{*}{Method} & DAIR-V2X      & OPV2V     & V2XSet   & V2V4Real   & Mean         \\
                        & AP@0.3/0.5 & AP@0.3/0.5 & AP@0.3/0.5 & AP@0.3/0.5 & AP@0.3/0.5 \\ \midrule
Baseline                    & 68.67/64.79 & 34.90/30.37 & 35.05/30.28 & \textbf{52.84}/31.63 & 47.86/39.27  \\
IBN-Net                 & 69.80/66.17 & 36.26/32.53 & 32.85/29.69 & 47.05/32.58 & 46.49/40.24  \\
MLDG$_a$                 & \textbf{71.61}/\textbf{66.84} & 36.02/30.95 & 36.50/31.11 & 52.18/32.27 & 49.08/40.29  \\
MLDG$_b$                   & 70.53/66.31 & 39.66/35.43 & 34.95/30.66 & 52.27/\textbf{33.42} & 49.35/41.45  \\
Ours                    & 70.26/66.31 & \textbf{40.94}/\textbf{36.54} & \textbf{40.82}/\textbf{36.20} & 52.52/33.05 & \textbf{51.13}/\textbf{43.03}  \\ \bottomrule
\end{tabular}}
\label{tab:dairv2x}
\end{table}

\begin{table*}[htb]
\centering
\caption{\textbf{Ablation study.} Effects of different components of our method on the OPV2V2ALL setting. We report the AP@0.3/0.5 and the improvement of each component (PG: Probabilistic Gate, PA: Point Augmentation). The first row is the result by the Baseline method.}
\resizebox{2.\columnwidth}{!}{%
\begin{tabular}{ccc|cc|cccccc|cc}
\hline
\multicolumn{2}{c}{CMAG}                                                   & CFC                      & \multicolumn{2}{c|}{OPV2V} & \multicolumn{2}{c}{V2XSet} & \multicolumn{2}{c}{V2V4Real} & \multicolumn{2}{c|}{DAIR-V2X} & \multicolumn{2}{c}{Mean} \\
 PG                       & PA                       &                           & AP@0.3       & AP@0.5      & AP@0.3       & AP@0.5      & AP@0.3        & AP@0.5       & AP@0.3        & AP@0.5        & AP@0.3     & AP@0.5    \\ \hline
                          &                           &                           & 90.57         & 90.25        & 85.13         & 84.39        & 39.96          & 38.74         & 41.44          & 36.79          & 64.27       & 62.54      \\ \hline
   \checkmark                        &                           &                           & 92.99(\textbf{\color{gray} +2.42}) & 92.47(\textbf{\color{gray} +2.22}) & \textbf{89.86(\textbf{\color{gray} +4.73})} & \textbf{88.63(\textbf{\color{gray} +4.24})} & 42.94(\textbf{\color{gray} +2.98}) & 41.17(\textbf{\color{gray} +2.43}) & 41.96(\textbf{\color{gray} +0.52}) & 37.04(\textbf{\color{gray} +0.24}) & 66.94(\textbf{\color{gray} +2.66}) & 64.83(\textbf{\color{gray} +2.28})      \\

 \checkmark & \checkmark &                           & \textbf{93.26(\textbf{\color{gray} +2.69})} & \textbf{92.60(\textbf{\color{gray} +2.34})} & 88.87(\textbf{\color{gray} +3.74}) & 87.24(\textbf{\color{gray} +2.85}) & 45.57(\textbf{\color{gray} +5.61}) & 43.17(\textbf{\color{gray} +4.43}) & 44.90(\textbf{\color{gray} +3.46}) & 39.65(\textbf{\color{gray} +2.86}) & 68.15(\textbf{\color{gray} +3.88}) & 65.66(\textbf{\color{gray} +3.12})      \\
\checkmark & \checkmark & \checkmark & 92.69(\textbf{\color{gray} +2.12}) & 92.03(\textbf{\color{gray} +1.78}) & 88.48(\textbf{\color{gray} +3.35}) & 87.42(\textbf{\color{gray} +3.03}) & \textbf{46.11(\textbf{\color{gray} +6.15})} & \textbf{44.17(\textbf{\color{gray} +5.43})} & \textbf{46.08(\textbf{\color{gray} +4.65})} & \textbf{41.13(\textbf{\color{gray} +4.33})} & \textbf{68.34(\textbf{\color{gray} +4.07})} & \textbf{66.19(\textbf{\color{gray} +3.64})}      \\ \hline
\end{tabular}}
\label{tab:ablation}
\vspace{-1.em}
\end{table*}

Table~\ref{tab:opv2v} shows the experiment performance trained on OPV2V~\cite{xu2022opv2v}. The Baseline method without any DG methods gain an  outstanding performance on source domain. However, it suffers from significant drops when deploying to other domains. While for OPV2V and V2XSet, their both agent number distribution and cooperative LiDAR setup are very similar as mentioned at Section~\ref{sec:introduction}. As a result, the drop for generalizations between these two datasets is small. 
On the contrary, the drops from OPV2V to V2V4Real and DAIR-V2X are much more pronounced, since the cooperative gaps are huge. After applying DG methods such as INB-Net~\cite{pan2018two} and MLDG$_a$~\cite{kim2023single}, the performance shows a slight improvement, due to the feature-level generalization capacity. Nevertheless, MLDG$_b$ may lead to some drops in performance, which might be caused by the lack of a more refined design tailored for cooperative perception. Notably, Our method almost achieves optimal performance across multiple domains, the AP@0.3/0.5 are improved by 2.12\%/1.78\% on OPV2V, 3.35\%/3.03\% on V2XSet, 6.15\%/5.43\% on V2V4Real and 4.65\%/4.33\% on DAIR-V2X compared with Baseline method. Our method gain mean AP@0.3/0.5 improvement of 4.07\%/3.65\% with Baseline, 2.12\%/2.40\% with IBN-Net, 2.05\%/2.00\% with MLDG$_a$ and 6.49\%/6.79\% with MLDG$_b$. 

Table~\ref{tab:v2xset}, Table~\ref{tab:v2v4real} and Table~\ref{tab:dairv2x} show the experiment performances trained on V2XSet~\cite{xu2022v2x}, V2V4Real~\cite{xu2023v2v4real} and DAIR-V2X~\cite{yu2022dair}, respectively. From the perspective of mean AP, our method demonstrates optimal generalization performance across various settings, underscoring its overall superiority.

\subsection{Ablation Study.}
To assess the contribution of each component in our proposed method, we conducted ablation studies using OPV2V as the source domain, referred to  the OPV2V2ALL setting. In Table~\ref{tab:ablation}, we present the AP@0.3/0.5 results as each component is incrementally added. The results demonstrate that all components of our method play a crucial role in enhancing the generalization ability and improving the overall cooperative perception performance. For our method, Table~\ref{tab:ablation} displays that the proposed CMAG contributes to the main improvement while the proposed CFC is able to further promote the generalization performance after CMAG.

\subsection{Qualitative Comparisons.}

In Fig.~\ref{fig:det_vis}, we visualize the qualitative results of the Baseline, IBN-Net~\cite{pan2018two} and our method when the model is trained on OPV2V and generalized to all datasets, and the detection errors are highlighted using purple arrows for better observation. All methods show great detection accuracy in OPV2V, but the performance varies when generalized to unseen domains,  illustrating the importance of generalization ability in V2X cooperative perception. The detection visualization highlights that our method has less detection error in unseen domains, which proves the excellent domain generalization performance of our method.

\section{Conclusions}\label{Sec:Conclusions}
Existing LiDAR-based V2X cooperative perception methods experience performance degradation when deployed in unseen domains. In this paper, we propose a novel domain generalization approach for LiDAR-based V2X cooperative perception (V2X-DG) to ensure robust performance in both unseen and source domains. Based on the observation and analysis of cooperative agent number distribution gap and cooperative agent type gap, we propose Cooperative Mixup Augmentation based Generalization (CMAG) to improve the model generalization capability by  simulating unseen cooperative cases. Furthermore, we also propose Cooperation Feature Consistency (CFC) as regularization of the robust generalized feature representation learning. The experimental results demonstrate the effectiveness of our proposed method. This research represents a significant advancement in V2X cooperative perception, enabling robust performance from the source domain to unseen domains.

\bibliographystyle{IEEEtran}
\bibliography{baolu}

\end{document}